\documentclass[conference]{IEEEtran}
\IEEEoverridecommandlockouts
\usepackage{cite}
\usepackage{amsmath,amssymb,amsfonts}
\usepackage{algorithmic}
\usepackage{graphicx}
\usepackage{hyperref}
\usepackage{textcomp}
\usepackage{float}
\usepackage{xcolor}
\def\BibTeX{{\rm B\kern-.05em{\sc i\kern-.025em b}\kern-.08em
    T\kern-.1667em\lower.7ex\hbox{E}\kern-.125emX}}

\makeatletter
\newcommand{\linebreakand}{%
  \end{@IEEEauthorhalign}
  \hfill\mbox{}\par
  \mbox{}\hfill\begin{@IEEEauthorhalign}
}
\makeatother

\begin{document}

\title{Plagiarism Detection in the Bengali Language: A Text Similarity-Based Approach \\
}

\author{\IEEEauthorblockN{Satyajit Ghosh 
\href{https://orcid.org/0000-0003-2791-5780}{\includegraphics[scale=0.1]{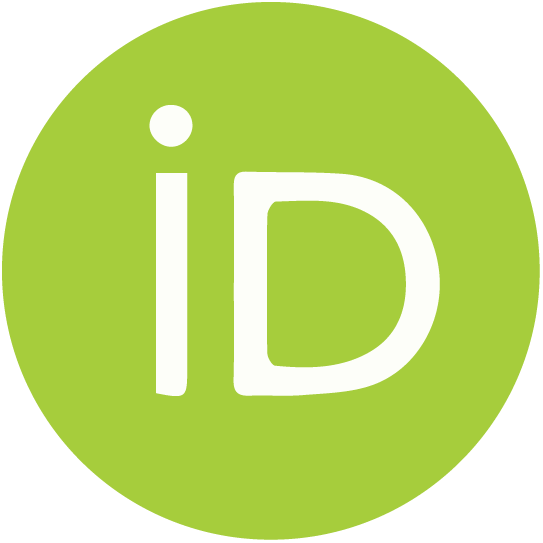}}}
\IEEEauthorblockA{\textit{Department of Computer Science \& Engineering} \\
\textit{Adamas University}\\
Kolkata, India \\
satyajit.ghosh@stu.adamasuniversity.ac.in}
\and
\IEEEauthorblockN{Aniruddha Ghosh}
\IEEEauthorblockA{\textit{Department of Computer Science \& Engineering} \\
\textit{SRM University}\\
Chennai, India \\
ag7434@srmist.edu.in}
\linebreakand
\IEEEauthorblockN{Bittaswer Ghosh}
\IEEEauthorblockA{\textit{Department of Computer Science \& Engineering} \\
\textit{Adamas University}\\
Kolkata, India \\
bittaswer.ghosh@stu.adamasuniversity.ac.in}
\and
\IEEEauthorblockN{Abhishek Roy}
\IEEEauthorblockA{\textit{Department of Computer Science \& Engineering} \\
\textit{Adamas University}\\
Kolkata, India \\
abhishek.roy@adamasuniversity.ac.in }
}

\maketitle

\begin{abstract}
Plagiarism means taking another person’s work and not giving any credit to them for it. Plagiarism is one of the most serious problems in academia and among researchers. Even though there are multiple tools available to detect plagiarism in a document but most of them are domain-specific and designed to work in English texts, but plagiarism is not limited to a single language only. Bengali is the most widely spoken language of Bangladesh and the second most spoken language in India with 300 million native speakers and 37 million second-language speakers. Plagiarism detection requires a large corpus for comparison. Bengali Literature has a history of 1300 years. Hence most Bengali Literature books are not yet digitalized properly. As there was no such corpus present for our purpose so we have collected Bengali Literature books from the National Digital Library of India and with a comprehensive methodology extracted texts from it and constructed our corpus. Our experimental results find out average accuracy between 72.10 \% - 79.89 \% in text extraction using OCR. Levenshtein Distance algorithm is used for determining Plagiarism. We have built a web application for end-user and successfully tested it for Plagiarism detection in Bengali texts. In future, we aim to construct a corpus with more books for more accurate detection.
\end{abstract}


\section{Introduction}
Cambridge advanced learner’s dictionary defines Plagiarism as “The process or practice of using another person’s ideas or work and pretending that it is your own” \cite{1}. As per the survey conducted by Josephson Institute Center for Youth Ethics, it has been found that 59\% of high school students admitted cheating and every one out of three students use the internet to plagiarize their assignments \cite{2}. Online learning and tests have seen exponential growth in the context of the COVID-19 pandemic. The quality of assignments should be checked properly by teachers to minimize plagiarism at an early stage \cite{3}. It is a time consuming and tedious job to check every paper for plagiarism. To help the teachers, evaluators, and researchers there are multiple Plagiarism detecting tools available in the market. These tools are mostly domain-specific and designed for detecting plagiarism in English texts. They are either incapable or inefficient for plagiarism detection in other languages, especially in Indian regional languages. Bengali is one such language. It is developed for roughly 1300 years and the timeline of Bengali literature is divided into three phases – ancient, medieval, and modern \cite{4}. Thus, Plagiarism detection in Bengali requires a corpus consisting of many old and new literature. The authors built one corpus with a small subset of Bengali Literature. Further, they performed Plagiarism detection using it to analyze its performance and accuracy.
The rest of the paper is organized as follows: Section 2 presents the proposed methodology and implementation for corpus creation. In Section 3 we discuss Plagiarism detection using our proposed algorithm and implementation of it using different tools and technologies. Section 4 reports the experimental results and performance of our detection tool. Finally, Section 5 concludes this paper.

\section{CORPUS CREATION}
A corpus is a collection of many processed, ordered, and selected texts. A text corpus is used for data mining \cite{5}, natural language processing \cite{6}, emotion analysis \cite{7}, Plagiarism detection \cite{8} and many more things. Our goal is to make a corpus that will be helpful for Plagiarism detection and in future can be used for other works as well.
\subsection{Proposed Methodology}
The proposed methodology is a step–by–step process for the corpus generation of Bengali Literature. Below is the pictorial representation of the proposed methodology.
\begin{figure}[h]
\caption{Proposed methodology for corpus creation}
\includegraphics[width=7cm]{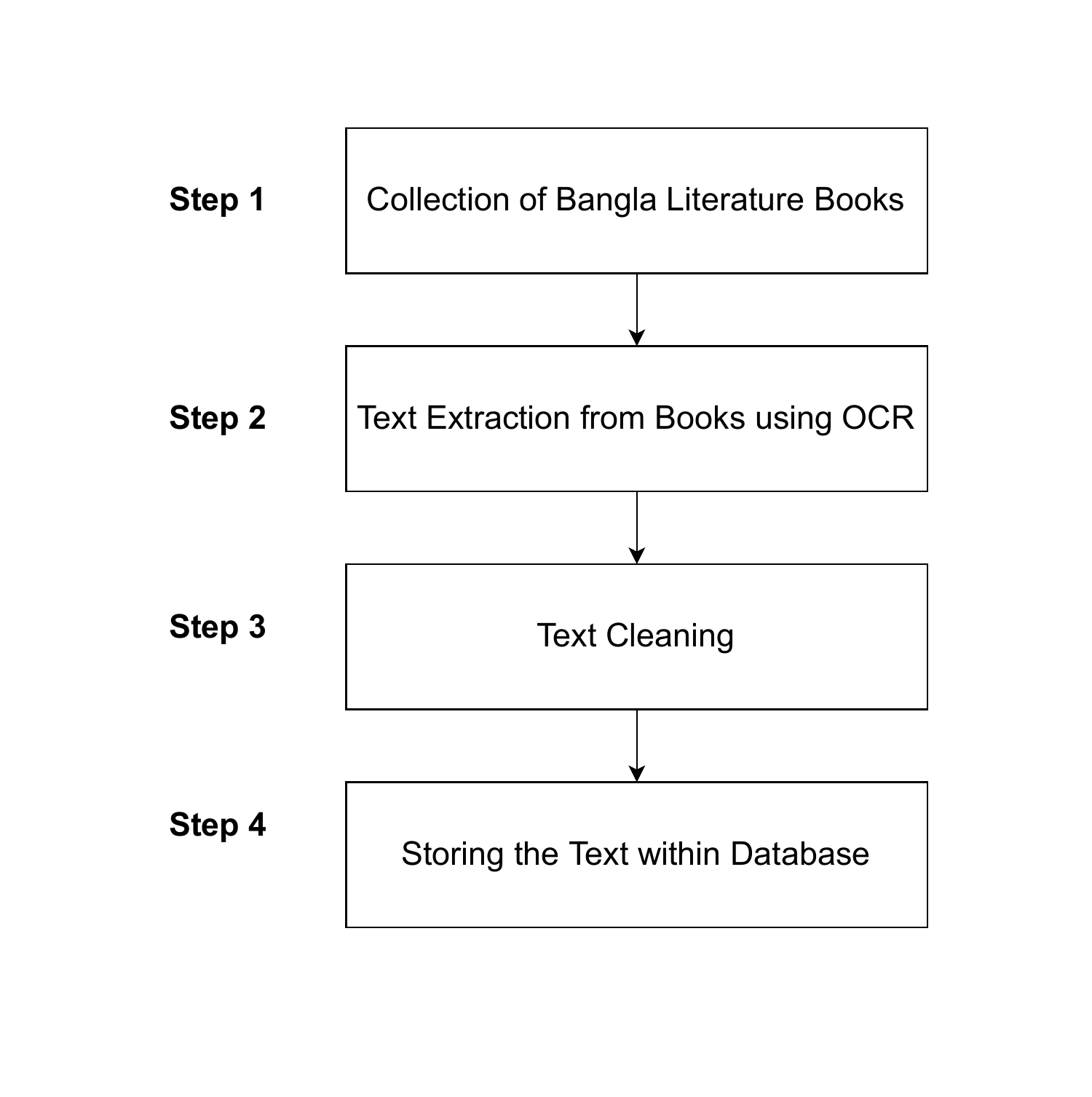}
\centering
\end{figure}
Step 1 includes the collection of Bengali literature books. National Digital Library of India at the time of writing this paper had more than 400 books on this topic from various sources. We have collected 200 books for the generation of the corpus. This includes both old and new books of Bengali Literature. Most of the books are present in the library are scanned PDF copies of physical books.
In Step 2 we have used Optical Character Recognition or OCR technology to extract the Bengali texts from the PDF files. Step 3 includes text cleanup as while extracting the texts from the books it has been found that invalid characters, non-Bengali characters, and whitespaces are coming out. So, we have removed all those from the texts and in Step 4 we have stored the cleaned texts in the database.
\subsection{Implementation}
\subsubsection{Required Tools and Technologies}
To generate the corpus, we need some tools and programs they are as follows:
\begin{itemize}
\item	Python3
\item	Jupyter notebook
\item	Tesseract OCR Engine
\item	SQLite3
\item	Microsoft Excel
\item	Pandas
\item	OpenCV
\end{itemize}
\subsubsection{Preparing Raw Data}
As a first step, we have collected Bengali Literature books from the National Digital Library of India and stored them in a directory after renaming them with unique identifiers and storing their other details in an Excel file.
\subsubsection{Performing OCR using Tesseract}
Tesseract is an open-source OCR engine that can be trained to recognize any language and it supports more than 100 languages out of the box \cite{9}. Bengali is also one such language that is supported by Tesseract out of the box using pre-trained models. Tesseract takes images as input for OCR operation, so we have converted the PDF files into images using a small Python program and stored the images on the hard disk on respective folders. Then we have provided the images as input to the OCR engine using OpenCV. Other image processing steps like Rescaling, Binarization, Noise Removal etc. are done by Tesseract internally \cite{10}, so we do not have to perform them.
\subsubsection{Text Cleanup}
After observing the extracted texts by Tesseract, it has been found that the texts contain whitespaces and invalid characters. As Bengali characters are present in the range of U+0980 to U+09FF in the Unicode block, so we have used a regular expression to remove any characters outside this range and the whitespaces.
\subsubsection{Storage}
To store the corpus, we have used the SQLite3 database. SQLite is a lightweight disk-based database, and it does not require a separate server process \cite{11}. After the clean-up process, the extracted texts are stored within the database. Next, a small Python program is used to store the other details of the book from the excel file to the SQLite database using Pandas library.
\begin{figure}[H]
\caption{ER Diagram of Database}
\includegraphics[width=7cm,trim=0cm 2cm 0cm 1cm,clip]{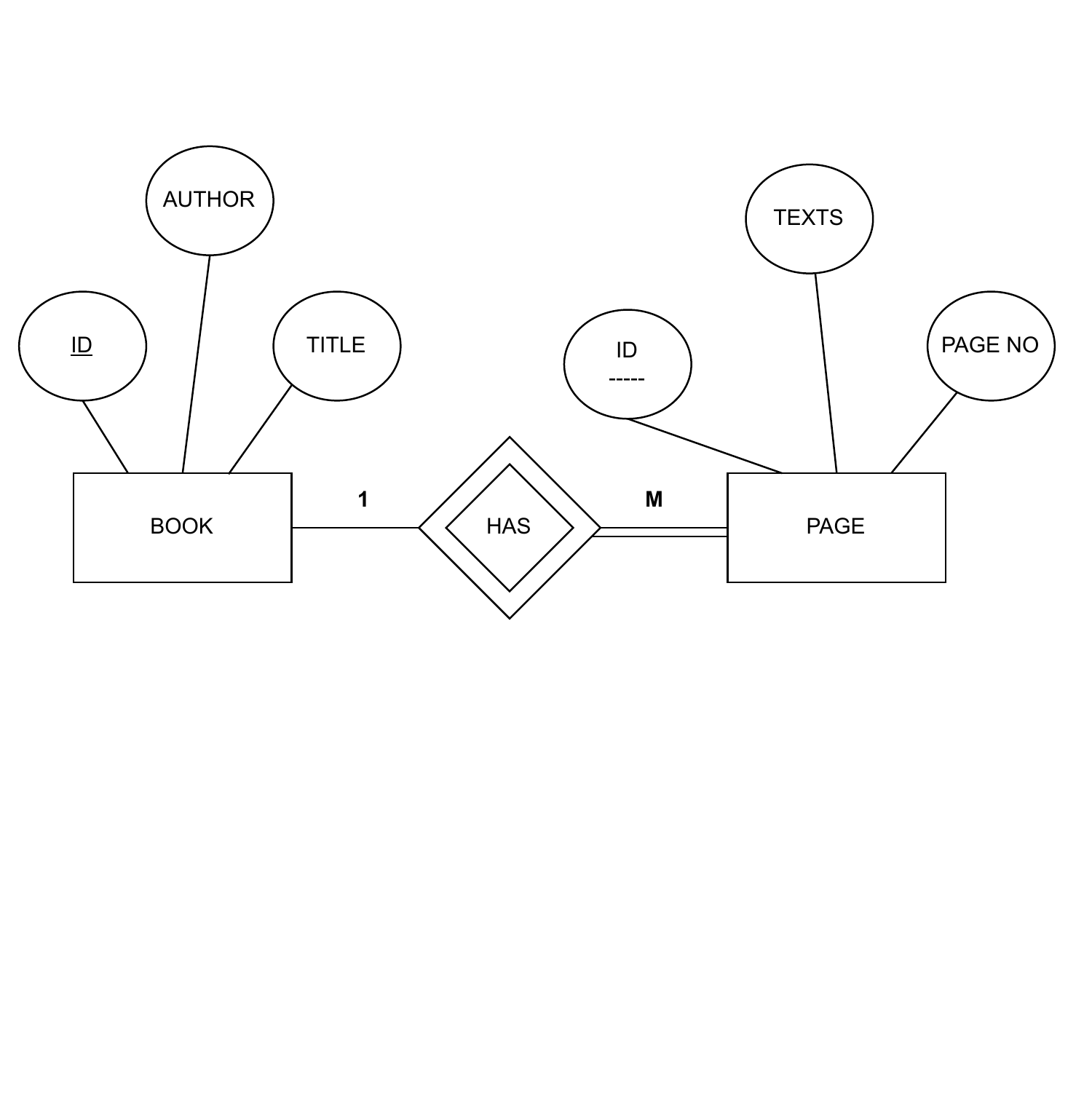}
\centering
\end{figure}
\subsection{Accuracy }
Accuracy is the quality of being correct and precise. It is very important to determine the accuracy of OCR operation for a good quality corpus. We have found that our collection can be grouped into two categories. The first category consists of old books where the images of scanned pages have noise, black marks on the border and unclear texts. We call it as “Dirty” category. The second category consists of the recent books where the images of scanned pages are mostly clear and without any noise. We call it as “Clean” category.
\begin{figure}[h]
\caption{OCR output on “Dirty” books}
\includegraphics[width=7cm]{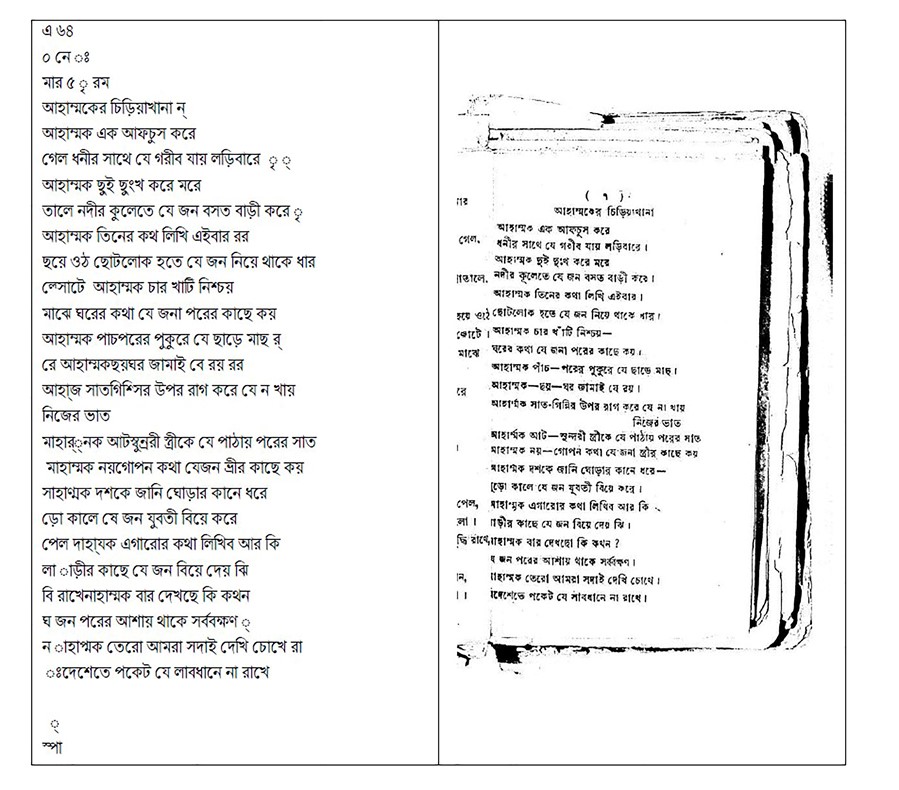}
\centering
\end{figure}
Fig.3. shows the extracted texts from a page where the text is not clear and has noise and an uneven border. We have selected random pages from the collection and found that we are having average accuracy of 72.10 \% from this category.
\begin{figure}[h]
\caption{OCR output on “Clean” books}
\includegraphics[width=8cm]{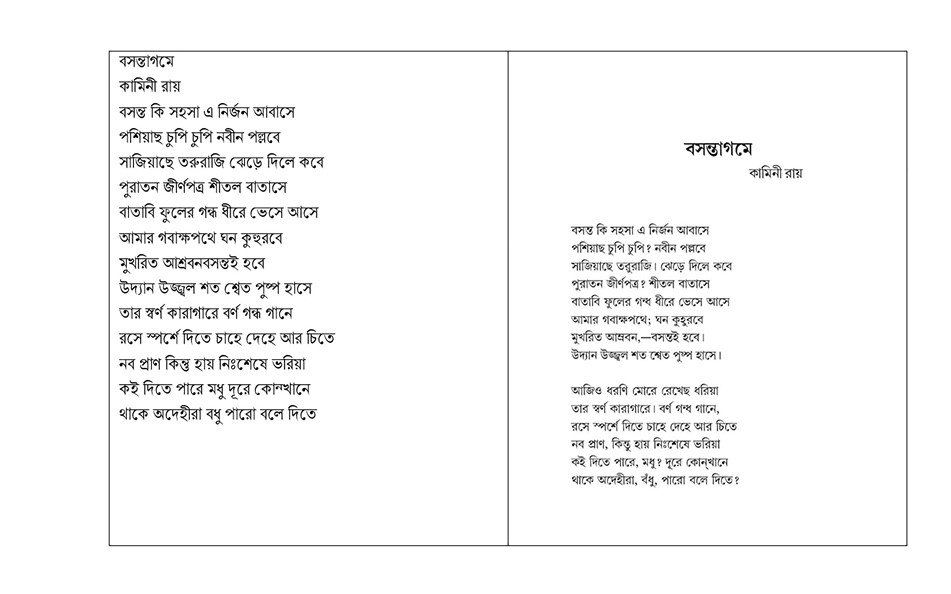}
\centering
\end{figure}
Fig.4. shows the extracted texts from a page where the text is clear and has no noise. We have selected random pages from the collection and found that we are having average accuracy of 79.89 \% from this category. 
For determining the accuracy, we have used Levenshtein Distance Algorithm. We have provided manually entered actual text of the page and the text which is extracted by the OCR engine to the algorithm. The algorithm checks for the similarity between two texts. We have discussed this algorithm in detail in the 3.1 section of this paper.
\section{PLAGIARISM DETECTION}
If one person represents another person’s work as their original work, then it is considered plagiarism. Plagiarism generally is not a crime but in academia and industry, it is an ethical offence \cite{12}. The exponential growth in digital resources increases the possibility of plagiarism. Plagiarism can appear in multiple ways like claiming another person's work as own or using another person's work without giving credit \cite{13}. We have Citation-Based, Semantic-Based, Cross Language-Based, Structural-Based and Character-Based methods for detecting Plagiarism in a text \cite{14}. Our proposed algorithm works based on text similarity between two documents.
\subsection{Proposed Algorithm}
The Levenshtein distance (also known as Edit Distance) is an algorithm that is used to measure the minimum number of edits required for changing one string to another using only three operations they are Insertion, Removal or Replacement of character. Levenshtein Distance algorithm can be implemented with a recursive solution or dynamic programming. The time and space complexity of this algorithm when implemented using dynamic programming is O (m * n) \cite{15,16}.

\begin{table}[]
\caption{Levenshtein Distance Algorithm \cite{17}}
\label{tab:my-table}
\resizebox{\columnwidth}{!}{%
\begin{tabular}{|l|l|}
\hline
Step & Description                                                                                                                                                                                                                                                                                                                    \\ \hline
1    & \begin{tabular}[c]{@{}l@{}}Set n to be the length of s.\\    Set m to be the length of t.\\    If n = 0, return m and exit.\\    If m = 0, return n and exit.\\    Construct a matrix containing 0..m rows and 0..n columns.\end{tabular}                                                                                      \\ \hline
2    & \begin{tabular}[c]{@{}l@{}}Initialize the first row to 0..n.\\    Initialize the first column to 0..m.\end{tabular}                                                                                                                                                                                                            \\ \hline
3    & Examine each character of s (i from 1 to n).                                                                                                                                                                                                                                                                                   \\ \hline
4    & Examine each character of t (j from 1 to m).                                                                                                                                                                                                                                                                                   \\ \hline
5    & \begin{tabular}[c]{@{}l@{}}If s{[}i{]} equals t{[}j{]}, the cost is 0.\\    If s{[}i{]} doesn't equal t{[}j{]}, the cost is 1.\end{tabular}                                                                                                                                                                                    \\ \hline
6    & \begin{tabular}[c]{@{}l@{}}Set cell d{[}i,j{]} of the matrix equal to the   minimum of:\\    a. The cell immediately above plus 1: d{[}i-1,j{]} + 1.\\    b. The cell immediately to the left plus 1: d{[}i,j-1{]} + 1.\\    c. The cell diagonally above and to the left plus the cost: d{[}i-1,j-1{]} +   cost.\end{tabular} \\ \hline
7    & After the iteration steps (3, 4, 5, 6) are   complete, the distance is found in cell d{[}n,m{]}.                                                                                                                                                                                                                               \\ \hline
\end{tabular}%
}
\end{table}

The Levenshtein algorithm score is inversely proportional to the similarity of two strings. Let us suppose we have two strings S1 and S2 having a length of M and N respectively and their Levenshtein algorithm score is denoted by DIFF then similarity calculation formula will be \cite{18}:
\begin{equation}
	Sim(S1,S2) = 1-\frac{DIFF}{max(M,N)} 
\end{equation}
\subsection{Implementation}
Our plagiarism checker application takes the user input. Then it tokenizes the user input and removes the stopwords. The same steps are followed for texts in the database. This helps us to reduce the time taken by the algorithm and improves its accuracy. After that, our algorithm compares the user input with every page of every book on the data-base and stores the similarity scores. After completing comparisons, it fetches the information about the pages in which the similarity scores are highest and met the threshold limit to consider it as plagiarism. In the end, it displays the result to the user along with book title, author name, page number and similarity score.
\begin{figure}[H]
\caption{Flowchart of Application}
\includegraphics[width=7cm]{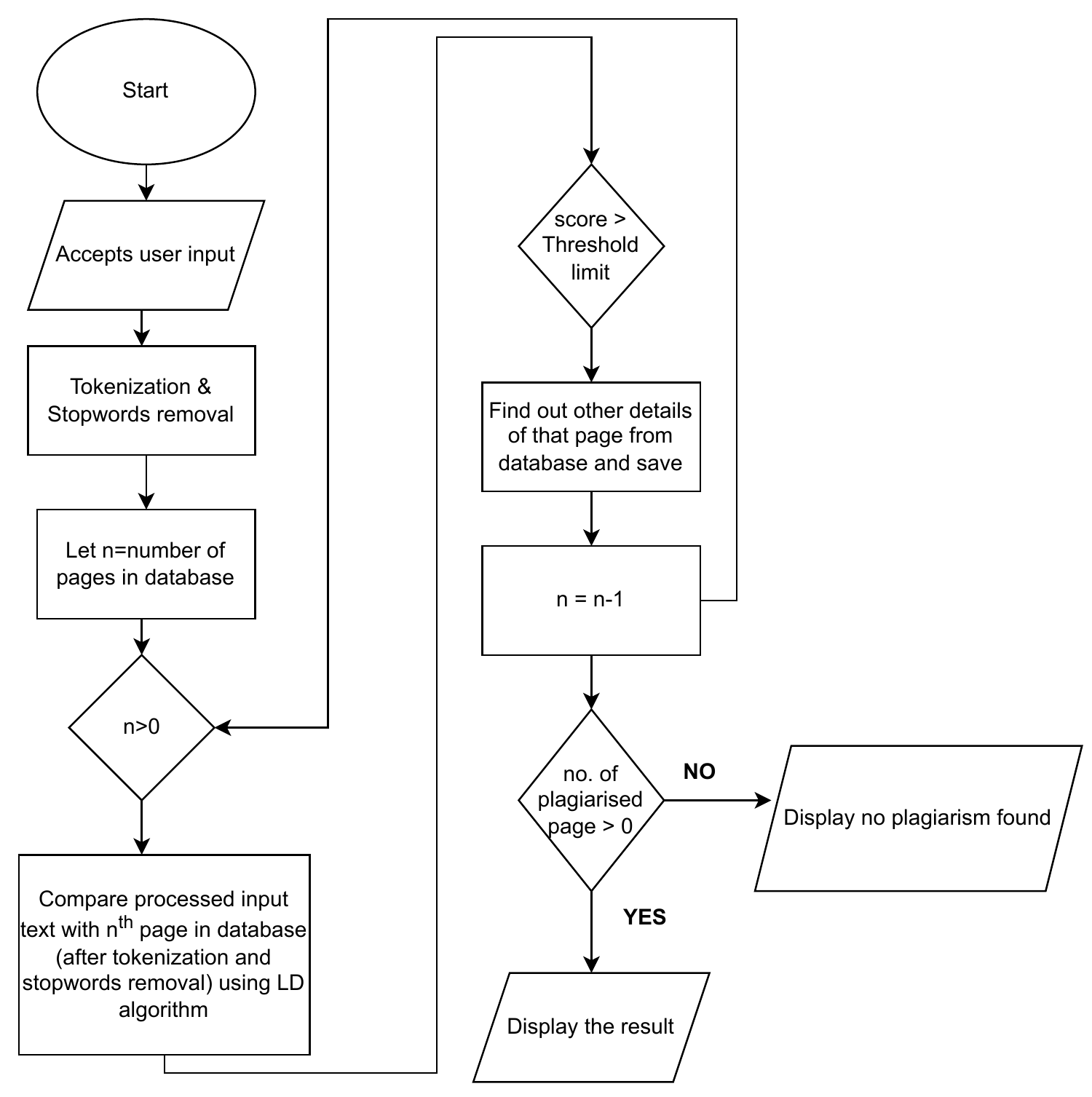}
\centering
\end{figure}
To implement the Bengali Plagiarism Checker, we have used Flask. Flask is a web framework that helps us to make web applications with a Python backend. The Flask application communicates with the SQLite database which we have prepared during corpus generation and serves the end-user.
\begin{figure}[H]
\caption{Web application for Plagiarism Detection on Bengali Literature}
\includegraphics[width=7cm]{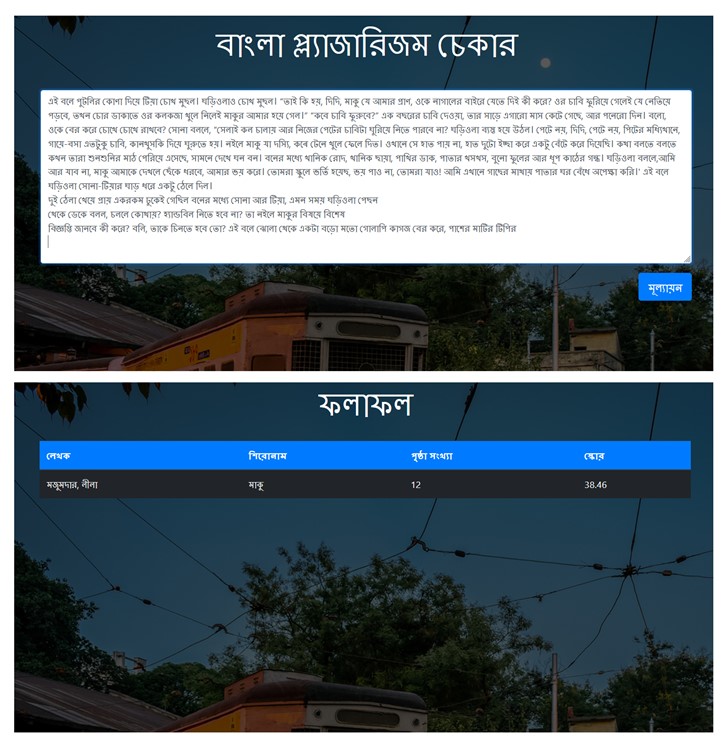}
\centering
\end{figure}
\section{EXPERIMENTAL RESULT AND PERFORMANCE ANALYSIS}
Here, we have provided texts as input. Then we have calculated the similarity score using our web application. We have observed our application can successfully find out the correct book title, author name and page number and similarity scores.
From our observations, we have found that in both the categories our Plagiarism detection is working properly except on a few “Dirty” books and the similarity score of 20 and above can confirm Plagiarism. The accuracy of the Plagiarism detection is proportional to the amount of text we give as input. More the text, the better the detection.
\section{CONCLUSION AND FUTURE WORK}
Plagiarism is one of the most serious problems faced by researchers and evaluators. E-learning and blended mode of teaching increase the probability of plagiarism by many times. This paper presented a comprehensive methodology for determining plagiarism in Bengali texts coupled with the corpus generation process for it. The same methodology can be applied to other regional languages for plagiarism detection. In this research, the Levenshtein Distance algorithm is successfully implemented on a web application for serving end-users requests to determine plagiarism in Bengali literature. All the source codes are available on our GitHub repository\footnote{Repository Link: \url
{ https://github.com/Bengali-Plagiarism-Tool}}. 
The volume of Bengali literature is huge. The collection of a greater number of books will increase the capability of the detection tool. In future, a web scraper can be built to collect a greater number of books. Though the proposed algorithm works well the time complexity of it is high and takes over a minute to provide the results. More efficient use of data structure and modified version of the proposed algorithm may reduce the detection time.  
\bibliographystyle{ieeetr}  
\bibliography{references}

\begin{thebibliography}{10}

\bibitem{1}
E.~Walter, {\em Cambridge Advanced Learner's Dictionary with CD-ROM}.
\newblock Cambridge university press, 2008.

\bibitem{2}
``2012 report card,'' Nov 2021.

\bibitem{3}
E.~Marais, U.~Minnaar, and D.~Argles, ``Plagiarism in e-learning systems:
  Identifying and solving the problem for practical assignments,'' in {\em
  Sixth IEEE International Conference on Advanced Learning Technologies
  (ICALT'06)}, pp.~822--824, IEEE, 2006.

\bibitem{4}
``Bengali literature,'' Jun 2022.

\bibitem{5}
Q.~Li, S.~Li, S.~Zhang, J.~Hu, and J.~Hu, ``A review of text corpus-based
  tourism big data mining,'' {\em Applied Sciences}, vol.~9, no.~16, p.~3300,
  2019.

\bibitem{6}
V.~Liu and J.~R. Curran, ``Web text corpus for natural language processing,''
  in {\em 11th Conference of the European Chapter of the Association for
  Computational Linguistics}, pp.~233--240, 2006.

\bibitem{7}
Y.~Kumar, D.~Mahata, S.~Aggarwal, A.~Chugh, R.~Maheshwari, and R.~R. Shah,
  ``Bhaav-a text corpus for emotion analysis from hindi stories,'' {\em arXiv
  preprint arXiv:1910.04073}, 2019.

\bibitem{8}
R.~R. Naik, M.~B. Landge, and C.~N. Mahender, ``Development of marathi text
  corpus for plagiarism detection in the marathi language,'' {\em corpus},
  vol.~6, p.~340, 2011.

\bibitem{9}
Tesseract-Ocr, ``Tesseract-ocr/tesseract: Tesseract open source ocr engine
  (main repository).''

\bibitem{10}
``Improving the quality of the output.''

\bibitem{11}
``Sqlite3 - db-api 2.0 interface for sqlite databases¶.''

\bibitem{12}
S.~P. Green, ``Plagiarism, norms, and the limits of theft law: Some
  observations on the use of criminal sanctions in enforcing intellectual
  property rights,'' {\em Hastings LJ}, vol.~54, p.~167, 2002.

\bibitem{13}
H.~A. Chowdhury and D.~K. Bhattacharyya, ``Plagiarism: Taxonomy, tools and
  detection techniques,'' {\em arXiv preprint arXiv:1801.06323}, 2018.

\bibitem{14}
A.~H. Osman, N.~Salim, and A.~Abuobieda, ``Survey of text plagiarism
  detection,'' {\em Computer Engineering and Applications Journal}, vol.~1,
  no.~1, pp.~37--45, 2012.

\bibitem{15}
G.~Navarro, ``A guided tour to approximate string matching,'' {\em ACM
  computing surveys (CSUR)}, vol.~33, no.~1, pp.~31--88, 2001.

\bibitem{16}
S.~S. Skiena, {\em The algorithm design manual}, vol.~2.
\newblock Springer, 1998.

\bibitem{17}
M.~Gilleland, ``Merriam park software,'' {\em Levenshtein Distance, in Three
  Flavors}, p.~52.

\bibitem{18}
S.~Zhang, Y.~Hu, and G.~Bian, ``Research on string similarity algorithm based
  on levenshtein distance,'' in {\em 2017 IEEE 2nd Advanced Information
  Technology, Electronic and Automation Control Conference (IAEAC)},
  pp.~2247--2251, IEEE, 2017.

\end{thebibliography}
\end{document}